\documentclass[pdflatex,sn-mathphys-num]{sn-jnl}

\usepackage{graphicx}%
\usepackage{multirow}%
\usepackage{amsmath,amssymb,amsfonts}%
\usepackage{amsthm}%
\usepackage{mathrsfs}%
\usepackage[title]{appendix}%
\usepackage{xcolor}%
\usepackage{textcomp}%
\usepackage{manyfoot}%
\usepackage{booktabs}%
\usepackage{algorithm}%
\usepackage{algorithmicx}%
\usepackage{algpseudocode}%
\usepackage{listings}%
\usepackage{enumitem}

\theoremstyle{thmstyleone}%
\theoremstyle{thmstyletwo}%

\theoremstyle{thmstylethree}%

\begin{document}

\title{World model inspired sarcasm reasoning with large language model agents}


\author*[1,2]{\fnm{Keito} \sur{Inoshita}}\email{inosita.2865@gmail.com}

\author[3]{\fnm{Shinnosuke} \sur{Mizuno}}\email{mizuno-shinnosuke658@g.ecc.u-tokyo.ac.jp}

\affil*[1]{\orgdiv{Faculty of Business and Commerce}, \orgname{Kansai University}, \orgaddress{\street{3-3-35 Yamatecho}, \city{Suita}, \postcode{5648680}, \state{Osaka}, \country{Japan}}}

\affil[2]{\orgdiv{Data Science and AI Innovation Research Promotion Center}, \orgname{Shiga University}, \orgaddress{\street{1-1-1 Baba}, \city{Hikone}, \postcode{5228522}, \state{Shiga}, \country{Japan}}}

\affil[3]{\orgdiv{Faculty of Medicine}, \orgname{The University of Tokyo}, \orgaddress{\street{7-3-1 Hongo}, \city{Bunkyo}, \postcode{1138654}, \state{Tokyo}, \country{Japan}}}


\abstract{
Sarcasm understanding is a challenging problem in natural language processing, as it requires capturing the discrepancy between the surface meaning of an utterance and the speaker’s intentions as well as the surrounding social context. Although recent advances in deep learning and Large Language Models (LLMs) have substantially improved performance, most existing approaches still rely on black-box predictions of a single model, making it difficult to structurally explain the cognitive factors underlying sarcasm. Moreover, while sarcasm often emerges as a mismatch between semantic evaluation and normative expectations or intentions, frameworks that explicitly decompose and model these components remain limited.
In this work, we reformulate sarcasm understanding as a world model inspired reasoning process and propose World Model inspired SArcasm Reasoning (WM-SAR), which decomposes literal meaning, context, normative expectation, and intention into specialized LLM-based agents. The discrepancy between literal evaluation and normative expectation is explicitly quantified as a deterministic inconsistency score, and together with an intention score, these signals are integrated by a lightweight Logistic Regression model to infer the final sarcasm probability. This design leverages the reasoning capability of LLMs while maintaining an interpretable numerical decision structure.
Experiments on representative sarcasm detection benchmarks show that WM-SAR consistently outperforms existing deep learning and LLM-based methods. Ablation studies and case analyses further demonstrate that integrating semantic inconsistency and intention reasoning is essential for effective sarcasm detection, achieving both strong performance and high interpretability.
}


\maketitle

\section{Introduction}\label{sec1}

Sarcasm is a highly sophisticated pragmatic phenomenon in which an utterance that appears positive on the surface actually conveys negative or critical intent, creating a discrepancy between the literal meaning of the utterance and the speaker’s true intention \cite{1}. It widely appears in real-world texts such as online debates, social media posts, reviews, and dialogue systems, and directly affects the performance of many Natural Language Processing (NLP) tasks, including sentiment analysis, intent understanding, and dialogue response generation \cite{2,3,4}. However, understanding sarcasm requires more than interpreting literal evaluative meaning; it also involves reasoning about how the situation should normally be evaluated according to social norms, as well as inferring the speaker’s intentions and emotions through Theory of Mind (ToM) \cite{5}. As such, sarcasm goes far beyond a simple sentence classification problem and entails substantial cognitive complexity.

In recent years, the development of Pretrained Language Models (PLMs) such as BERT and RoBERTa, as well as Large Language Models (LLMs), has led to significant performance improvements in sarcasm detection \cite{6}. Task-specific deep learning models and zero-shot or prompt-based approaches using LLMs have achieved high accuracy; however, in most cases, sarcasm judgment is entrusted to a single black-box model. As a result, it remains structurally unclear which information the model relies on to identify sarcasm, and when errors occur, it is difficult to analyze whether the failure arises from semantics, context, or intention, leaving a fundamental problem of interpretability unresolved \cite{7}.

Furthermore, most existing studies treat sarcasm detection as a direct mapping from an input sentence to a label, without explicitly modeling how the multiple cognitive processes that humans are believed to engage in, literal interpretation, context inference, norm-based expectation, and intention reasoning, interact within the model. In particular, although the discrepancy between surface meaning and socially expected evaluation lies at the core of sarcasm understanding, there is still no established framework that captures this discrepancy itself in a numerical and structural manner and disentangles it from other factors. Meanwhile, in the field of computer vision, such discrepancies are addressed using world models \cite{8}. A world model reproduces the typical human perceptual process—observation $\rightarrow$ latent state $\rightarrow$ prediction $\rightarrow$ prediction error $\rightarrow$ decision—and this structure is potentially well suited for sarcasm understanding as well. Since sarcasm can be naturally defined as a mismatch between observation and prediction, it aligns with a world model structure in which prediction error is explicitly computed. This perspective enables structural tracing of “where the inversion occurs,” which is difficult for black-box classifiers. Therefore, to clarify what constitutes the essential cues for sarcasm and how they should be integrated, a world model inspired approach to sarcasm understanding is required.

In this work, we reinterpret sarcasm understanding as a world model inspired reasoning process consisting of observation, latent state inference, norm-based prediction, prediction error, and intention judgment, and implement it as a combination of multiple LLM agents and a lightweight learner, termed World Model inspired SArcasm Reasoning (WM-SAR). WM-SAR explicitly decomposes literal evaluation, context construction, norm-based expectation reasoning, computation of semantic inconsistency, and ToM-based intention reasoning into independent agents, and integrates their numerical signals in the final stage using Logistic Regression (LR) to achieve both high performance and interpretability. Through this design, we aim to leverage LLMs not merely as classifiers, but as modular components that collectively realize a world model inspired reasoning structure.

The contributions of this work are summarized as follows:
\begin{itemize}
\item[i)] A world model inspired conceptual framework for sarcasm understanding is introduced, which focuses on the relationship between the discrepancy of literal meaning and context-based normative expectation and the speaker’s intentional exploitation of this discrepancy, and theoretically organizes the cognitive structure of sarcasm as an integration of semantic inconsistency and intentional use.

\item[ii)] WM-SAR, a novel world model inspired reasoning framework, is proposed, which decomposes literal meaning, context, norm-based expectation, and intention reasoning into LLM agents and integrates them via deterministic difference computation and a lightweight LR, establishing a methodology that decomposes sarcasm judgment into interpretable numerical signals and natural language rationales.

\item[iii)] Extensive experiments on multiple benchmark datasets show that WM-SAR consistently outperforms existing deep learning and LLM based methods, and ablation analyses quantitatively verify that both semantic inconsistency and intention reasoning are indispensable for performance improvement.
\end{itemize}

The remainder of this paper is organized as follows. Section~2 reviews related work. Section~3 presents the details of WM-SAR. Section~4 reports experimental results, and Section~5 discusses the findings. Finally, Section~6 concludes the paper.

\section{Related work}\label{sec2}

\subsection{Neural networks for sarcasm detection}\label{subsec1}

Sarcasm detection has long been recognized as one of the most challenging problems in sentiment analysis. Early studies mainly relied on machine learning with handcrafted features, such as lexical cues, sentiment lexicons, and syntactic patterns, and showed that incorporating cognitive features, including readers' eye movement signals, could complement ambiguities that are difficult to capture from text alone \cite{9}. Analyses focusing on the discrepancy between positive expressions and negative emotions were also conducted on Twitter data by combining NLP and corpus based methods \cite{10}.

Subsequent work explored a variety of approaches, including reducing misclassification through refined feature engineering \cite{11}, multilingual sarcasm detection by introducing news context \cite{12}, classification of sarcasm types and algorithmic comparisons \cite{13}, and comprehensive studies of computational sarcasm ranging from rule based methods to machine learning \cite{14}. These studies suggest that sarcasm emerges as a discrepancy between surface meaning and latent emotions or situations. However, only limited efforts explicitly modeled this discrepancy structure inside the model.

With the advent of deep learning, distributed word representations such as Word2Vec and GloVe \cite{15} enabled Convolutional Neural Networks (CNNs) \cite{16}, Recurrent Neural Networks (RNNs), and Long Short-Term Memory (LSTM) networks \cite{17} to learn hierarchical and sequential features, leading to substantial improvements in sarcasm detection performance. To capture structural dependencies within sentences, graph neural network based approaches were also proposed \cite{18}. Focusing directly on incongruity in sarcasm, a dual branch architecture combined with evidential deep learning was introduced, demonstrating that discrepancy structures provide effective cues for sarcasm detection \cite{19}. Moreover, a framework based on multi scale convolution and contrastive feature alignment was proposed to handle feature consistency under imbalanced data settings \cite{20}. Despite their strong performance, these models mainly embed pragmatic elements such as implicit meaning and rhetorical intention into latent representations, making it difficult to explain which factors actually contribute to sarcasm judgments.

More recently, zero shot sarcasm detection using LLMs has attracted attention, and Generative Pre trained Transformer based models have been reported to achieve strong performance with minimal supervision \cite{21}. Furthermore, the introduction of intermediate reasoning through Chain of Thought (CoT) prompting \cite{22}, along with extensions such as Auto CoT \cite{23}, Tree of Thought, and Graph of Thought \cite{24}, has shown that exploring multiple reasoning paths can improve contextual understanding. Nevertheless, these LLM based approaches still ultimately rely on black box decisions made by a single model. While they enhance representation learning for sarcasm detection, they do not structurally separate and connect cognitive components such as meaning, context, normative expectation, and intention.

In summary, existing studies have succeeded in building high performance classifiers, but they do not provide sufficient answers to how the discrepancy between surface meaning and context dependent normative expectation should be understood as a combination of cognitive processes, nor what kind of framework can explain the internal structure in a human interpretable manner. In contrast, this work aims to reformulate sarcasm understanding as a world model inspired reasoning structure, explicitly identifying where discrepancies arise and what roles they play.

\subsection{Multi-agent systems for human imitation}\label{subsec2}

In complex reasoning tasks, multi-agent frameworks in which multiple specialized agents collaborate have recently attracted significant attention. Studies have reported improved diversity and accuracy through debate based reasoning in CAMEL \cite{25}, role assigned dialogue generation \cite{26}, and structured interaction for mathematical problem solving \cite{27}, demonstrating robustness beyond what can be achieved by a single model. In particular, frameworks such as CAMEL and AutoGen \cite{28} attempt to imitate human like reasoning processes through role division and interactive communication among agents. Structured reasoning, originating from CoT, has further evolved into Tree of Thought and Graph of Thought, providing mechanisms to explore and integrate multiple reasoning trajectories. These frameworks enable flexible thinking that does not depend on a single reasoning sequence and broaden applicability to tasks involving ambiguity.

Sarcasm understanding inherently requires the integration of multiple cognitive processes, including interpretation of literal meaning, completion of implicit context, assumption of social normative expectation, and inference of speaker intention and emotion \cite{29}. However, most previous sarcasm detection studies have treated these factors by embedding them into a single representation space, and very few models have explicitly decomposed the roles and relationships of each cognitive component. From this perspective, assigning roles such as meaning, context, normative expectation, and intention to different LLM agents and coordinating them to make a decision is a natural direction for imitating human cognitive processes. Nevertheless, existing multi-agent research has mainly focused on tasks such as mathematical reasoning or dialogue generation, and systematic applications to pragmatic phenomena like sarcasm understanding remain largely unexplored.

This work redesigns the multi-agent collaboration framework as a cognitive decomposition structure specialized for sarcasm understanding, and further aligns it with the world model inspired flow of observation to latent state to prediction to prediction error to decision. By doing so, we take a position that addresses a question left implicit in previous studies: which discrepancies among cognitive components give rise to sarcasm. In this sense, this work provides a new connection between multi-agent research and sarcasm detection research.

\section{Multi-agent world model inspired reasoning}
\label{sec:multiagent}

This study formulates sarcasm understanding as the detection of semantic inversion arising from world model inspired reasoning, and proposes an LLM based multi-agent reasoning framework that mimics its underlying cognitive process. Conventional sarcasm detection models mostly adopt a black box design that directly maps an input sentence to a label by learning a classifier over features obtained from a single encoder. Such designs make it difficult to structurally explain why a model judges an utterance as sarcastic, treat contextual information and social norms only implicitly, and do not allow analysis from the perspectives of world model inspired reasoning or ToM, which are essential for sarcasm understanding.

In contrast, the proposed framework explicitly implements the following five cognitive components that are considered essential for sarcasm understanding as independent agents: i) literal semantics, ii) context inference, iii) norm-based expectation modeling, iv) prediction error detection between observation and expectation, and v) ToM reasoning for speaker intention. Each agent performs inference on the input text in parallel, and their outputs are aggregated by the final integrator, referred to as the Sarcasm Arbiter.

A key design choice of this study is to perform the final integration not by an LLM, as commonly adopted in many multi-agent approaches, but by a lightweight LR. This design preserves a structure in which interpretable signals produced by agents are integrated by an explainable decision layer, rather than delegating the final judgment to black box free-form reasoning of an LLM. Moreover, since LR is parameterized by explicit coefficients, it enables quantitative analysis of how each signal contributes to the final decision. At the same time, because the inputs are restricted to low-dimensional interpretable features and no high-capacity additional classifier is introduced, a one-to-one correspondence with the world model structure of observation $\rightarrow$ latent state $\rightarrow$ prediction $\rightarrow$ prediction error $\rightarrow$ decision can be maintained.

\subsection{Problem formulation}
\label{sec:problem}

The input is a single text $u$, and the dataset assigns a sarcasm label $\mathrm{sarcasm}(u) \in \{0,1\}$ to each text. The objective of this study is to estimate the probability $P(\mathrm{sarcasm}(u)=1 \mid u)$ that a given text $u$ is sarcastic.

In psycholinguistics and pragmatics, sarcasm is characterized as an inversion between literal meaning and norm-based expectation. This notion is formalized in this study as
\begin{equation}
\mathrm{sarcasm}(u)=1
\quad\text{if}\quad
\operatorname{sgn}\!\big(M_{\mathrm{literal}}(u)\big)
\neq
\operatorname{sgn}\!\big(E_{\mathrm{norm}}(C(u))\big),
\label{eq:sign_flip}
\end{equation}
where $M_{\mathrm{literal}}(u)$ denotes the semantic valence of the literal meaning of text $u$, $C(u)$ represents the contextual latent state describing the background situation in which the text could occur, and $E_{\mathrm{norm}}(C(u))$ denotes the norm-driven expected valence under general social norms for that situation.

The sign function $\operatorname{sgn}(\cdot)$ is defined to allow a neutral region as
\begin{equation}
\operatorname{sgn}(x)=
\begin{cases}
+1, & x > \epsilon,\\
0, & |x| \le \epsilon,\\
-1, & x < -\epsilon,
\end{cases}
\label{eq:sgn}
\end{equation}
with $\epsilon > 0$. In the experiments, $\epsilon$ is fixed to $0.05$ in order to prevent spurious sign flips caused by small fluctuations in literal valence and norm-based expectation. With this formulation, sarcasm is described as a state in which the norm-driven expected valence $E_{\mathrm{norm}}(C(u))$ predicted by the world model is inverted with respect to the observed literal valence $M_{\mathrm{literal}}(u)$. The proposed framework aims to explicitly reproduce this reasoning structure through cooperative inference of LLM agents.
\subsection{Framework overview}
\label{sec:framework_overview}

\begin{figure}[t]
    \centering
    \includegraphics[width=\linewidth]{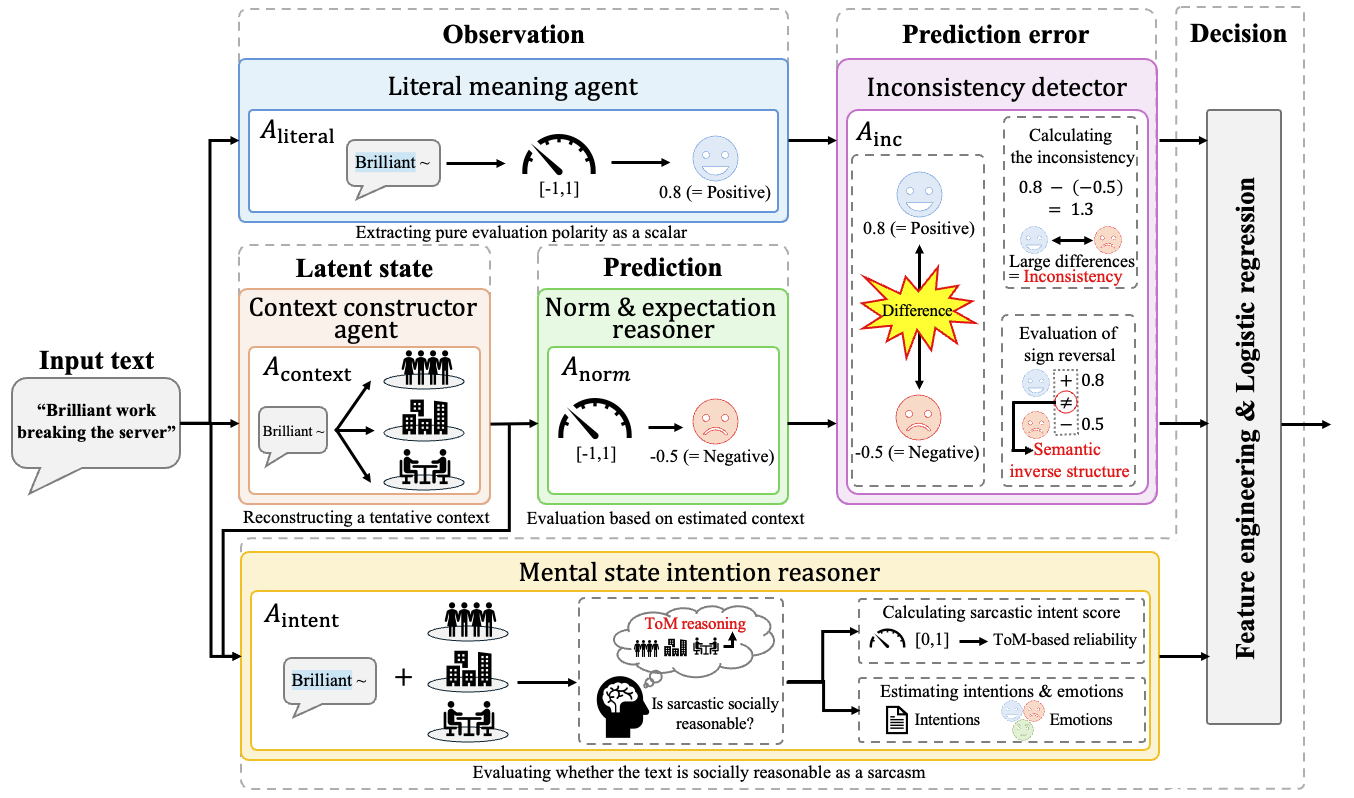}
    \caption{Framework overview of the WM-SAR.}
    \label{fig:framework}
\end{figure}

This study proposes WM-SAR, which reinterprets sarcasm understanding as a computational structure of world model inspired reasoning and implements it as a combination of a set of LLM agents and a lightweight learner. The overall architecture is illustrated in Fig.~\ref{fig:framework}. The framework explicitly decomposes the cognitive processes required for sarcasm understanding into five LLM agents. These agents are not merely implementation units, but are designed as role-sharing modules that correspond to the cognitive components underlying sarcasm understanding.

Specifically, the agent set is defined as
\begin{equation}
\mathcal{A}=
\{A_{\mathrm{literal}},A_{\mathrm{context}},A_{\mathrm{norm}},A_{\mathrm{inc}},A_{\mathrm{intent}}\}.
\end{equation}
Each agent $A_{\bullet}$ plays the following role:

\begin{enumerate}[label=\roman*)]
\item $A_{\mathrm{literal}}$ (Literal Meaning Agent):  
This agent extracts the evaluative polarity $M_{\mathrm{literal}}(u)$ from the literal meaning of the input text $u$. It plays the role of observation in the world model.

\item $A_{\mathrm{context}}$ (Context Constructor Agent):  
This agent constructs a representative background situation $C(u)$ in which the utterance could occur, as a hypothesis derived from the input text $u$. It plays the role of latent state inference in the world model.

\item $A_{\mathrm{norm}}$ (Norm and Expectation Reasoner):  
This agent estimates the norm-driven expected valence $E_{\mathrm{norm}}(C(u))$ based on social norms under the inferred context $C(u)$. It plays the role of prediction in the world model.

\item $A_{\mathrm{inc}}$ (Inconsistency Detector):  
This agent computes the inconsistency $D(u,C(u))$ and the sign discrepancy indicator $\mathrm{SD}(u,C(u))$ between the observation $M_{\mathrm{literal}}(u)$ and the prediction $E_{\mathrm{norm}}(C(u))$. It plays the role of prediction error computation in the world model.

\item $A_{\mathrm{intent}}$ (Mental State and Intention Reasoner):  
Based on the text $u$ and the context $C(u)$, this agent performs ToM reasoning to infer the speaker's intentions and emotions, and outputs the sarcasm intention alignment score $T_{\mathrm{sar}}(u,C(u))$. It provides the ToM perspective required for decision in the world model.
\end{enumerate}

Each agent $A_i \in \mathcal{A}$ receives the text $u$ and performs inference independently to produce its corresponding output $r_i$:
\begin{equation}
A_i:\ \text{input} \mapsto r_i.
\end{equation}
Here, $r_i$ is represented as a pair of a numerical scalar $z_i$ and a natural language rationale $\rho_i$:
\begin{equation}
r_i = (z_i, \rho_i).
\end{equation}
The numerical scalar $z_i$ is integrated by the downstream integrator, while the rationale $\rho_i$ is retained as auxiliary information for analysis, visualization, and error diagnosis.

In this study, the final judgment is not delegated to an LLM. Instead, the interpretable signals extracted by the agents are integrated by LR, and the sarcasm probability is estimated as
\begin{equation}
P(\mathrm{sarcasm}=1 \mid u) = F_{\mathrm{LR}}\big(\phi(u)\big),
\label{eq:lr_prob_overview}
\end{equation}
where $\phi(u)$ denotes the feature vector constructed from the agent outputs, typically including $|D(u,C(u))|$, $\mathrm{SD}(u,C(u))$, and $T_{\mathrm{sar}}(u,C(u))$. Integration by LR allows learning weight coefficients without introducing a high-capacity additional classifier, making it possible to quantitatively trace which signals contribute to the final decision, while maintaining compatibility between an interpretable world model structure and data-driven integration.

Furthermore, all agents in the framework are executed in parallel rather than sequentially. This design is not merely an implementation choice, but is motivated by cognitive science findings that multiple processes such as literal interpretation, situation inference, norm understanding, and intention inference proceed in a semi-parallel manner in human language understanding. Implementing each inference process as an independent agent enhances interpretability as a computational cognitive model. Since each agent outputs an independent numerical scalar, ablation is straightforward, enabling quantitative analysis of the contribution of each cognitive component to sarcasm understanding. In this sense, the framework serves not only as a high-performance model, but also as an experimental platform for cognitive contribution analysis.

Overall, the proposed method is formulated as WM-SAR, which combines modular decomposition into five cognitive agents, transparent integration by LR, and robust parallel inference.

\subsection{Component agents}
\label{sec:agents}

This subsection describes each agent that constitutes WM-SAR, clarifying its role and its correspondence to the world model structure.

\subsubsection{Literal meaning agent}
\label{sec:literal_agent}

The Literal Meaning Agent analyzes only the literal semantic content of the input text $u$ and extracts a pure evaluative polarity as a scalar, without considering context, social norms, speaker intention, or any pragmatic implication:
\begin{equation}
M_{\mathrm{literal}}(u) \in [-1,1].
\end{equation}
Here, $1$ denotes a superficially strong positive evaluation and $-1$ denotes a superficially strong negative evaluation. The agent is constrained by the output format so that the scalar is always mapped into the predefined range. This agent is intended to computationally approximate first-pass processing in sarcasm understanding, that is, the initial stage in which humans interpret an utterance literally. Therefore, it is designed to estimate polarity based solely on surface cues such as lexical choices and explicit evaluative expressions.

In conventional single-model approaches, literal evaluation is often assumed to be implicitly encoded within text embeddings, making it difficult to structurally explain which lexical or syntactic cues contribute to polarity judgments. Moreover, even when superficially positive expressions function sarcastically through contrast with context, separating surface evaluation from context-based evaluation has been difficult. To make this explicit, this study isolates literal evaluation as $M_{\mathrm{literal}}(u)$ and transparently connects it to the downstream norm-based expectation $E_{\mathrm{norm}}$ and inconsistency $D$, enabling direct handling of semantic inversion as a structure.

The output consists of the literal evaluation scalar $M_{\mathrm{literal}}(u)$ and a natural language explanation $\rho_{\mathrm{literal}}(u)$ that describes the rationale behind the judgment. Although $\rho_{\mathrm{literal}}(u)$ is not directly used for numerical integration, it is retained as auxiliary information for visualizing which surface cues contributed to the decision, and for subsequent error analysis and diagnosis.

Overall, this agent explicitly represents surface evaluation as an observation and, by separating it from intention reasoning, provides a foundation for structurally capturing the gap that is characteristic of sarcasm. In this framework, it plays the role of observation in the world model.

\subsubsection{Context constructor agent}
\label{sec:context_agent}

The Context Constructor Agent constructs, as a hypothesis, a typical and socially plausible background situation in which the input text $u$ could be uttered. This corresponds to latent state inference in the world model. Since many sarcasm datasets consist of single sentences without explicit dialogue history or external situations, models are forced to implicitly complement missing context in a black-box manner. In this study, general common sense and world knowledge embedded in LLMs are exploited to explicitly reconstruct a plausible virtual context $C(u)$ from a single text. The objective is not to recover the true context, but to obtain a representative context that is sufficient for norm-based expectation reasoning.

The background situation is represented as
\begin{equation}
C(u)=\{a^{(u)},\, s^{(u)},\, e^{(u)}\},
\end{equation}
where $a^{(u)}$ denotes the social relationship between the speaker and the listener, $s^{(u)}$ denotes the scene, and $e^{(u)}$ denotes the immediately preceding event. This decomposition does not aim at a complete description of the world state, but extracts the minimal sufficient factors required to derive normative evaluation. Importantly, this agent does not judge whether the utterance is sarcastic. Its role is strictly limited to generating a latent state hypothesis. This design preserves a reasoning structure in which context construction and evaluation are clearly separated for downstream norm reasoning and semantic inversion detection.

In addition, $C(u)$ generated by this agent is a hypothesized estimate and may contain errors and uncertainty. A mismatch in the context hypothesis can propagate to the inconsistency $D(u,C(u))$ through the norm-based expectation $E_{\mathrm{norm}}(C(u))$ computed by the Norm and Expectation Reasoner, thereby affecting the sensitivity of semantic inversion detection. In WM-SAR, however, the literal observation $M_{\mathrm{literal}}(u)$ and the ToM-based intention signal $T_{\mathrm{sar}}(u,C(u))$ are maintained as independent pathways, and LR integrates multiple signals at the final stage. This design prevents the final decision from relying excessively on a single context hypothesis. Nevertheless, for utterances with extremely strong context dependence or heavy implicit presuppositions, context estimation errors may affect performance, and this point is discussed later as a limitation.

Overall, this agent constructs a latent context in a self-contained manner from a single sentence and provides the latent state required for downstream norm reasoning. In this framework, it corresponds to latent state inference in the world model.

\subsubsection{Norm and expectation reasoner}
\label{sec:norm_agent}

The Norm and Expectation Reasoner infers how the situation represented by the context $C(u)$ estimated by the Context Constructor Agent should normally be evaluated, in light of generally shared social norms. This norm-driven expectation is represented as
\begin{equation}
E_{\mathrm{norm}}(C(u)) \in [-1,1].
\end{equation}
Here, $E_{\mathrm{norm}}(C(u))$ approximates a normative evaluation that is assumed to be shared by an average member of society, rather than a subjective opinion of a specific individual. This corresponds to state prediction in the world model.

Many sarcastic expressions arise when a superficially positive utterance is made toward a situation that would normally be evaluated as clearly negative from a social perspective. Therefore, this study emphasizes explicitly quantifying not the factual correctness of the situation, but the normative expectation of how it should be evaluated.

The output consists of the scalar value $E_{\mathrm{norm}}(C(u))$ and a natural language explanation $\rho_{\mathrm{norm}}(C(u))$ that describes the rationale behind the judgment. Although $\rho_{\mathrm{norm}}(C(u))$ is not directly used for numerical integration, it is retained as auxiliary information for visualizing which social expectations were referenced, and for subsequent error analysis and diagnosis.

By separating the roles such that the Context Constructor Agent generates a latent state hypothesis without evaluation and the Norm and Expectation Reasoner performs normative evaluation on top of it, the framework prevents implicit injection of evaluation during context construction and maintains a structure in which the difference between observation and expectation can be computed purely at a later stage.

Overall, this agent explicitly represents what would normally be thought in a given situation as a numerical value and provides the reference baseline for semantic inversion detection, serving as a core module for sarcasm understanding grounded in social common sense.

\subsubsection{Inconsistency detector}
\label{sec:inc_agent}

The Inconsistency Detector computes, as a deterministic formula, the inconsistency between the observation $M_{\mathrm{literal}}(u)$ output by the Literal Meaning Agent and the prediction $E_{\mathrm{norm}}(C(u))$ output by the Norm and Expectation Reasoner. In this study, the inconsistency is defined as
\begin{equation}
D(u,C(u)) = M_{\mathrm{literal}}(u) - E_{\mathrm{norm}}(C(u)).
\label{eq:diff}
\end{equation}
This directly corresponds to prediction error equals observation minus prediction in the world model, and reflects the fundamental stance of this study that sarcasm can be interpreted as prediction error.

This agent performs no additional language reasoning. By computing a simple difference, it visualizes the semantic gap in an interpretable manner and directly extracts the core structure of sarcasm as semantic inversion. Since $M_{\mathrm{literal}}$ and $E_{\mathrm{norm}}$ are normalized to the same scale $[-1,1]$, the difference yields a semantically consistent measure of inconsistency.

Furthermore, this study defines the presence of sign inversion as a structural condition by
\begin{equation}
\mathrm{SD}(u,C(u))
=
\mathbb{I}\!\left[
\operatorname{sgn}(M_{\mathrm{literal}}(u))
\neq
\operatorname{sgn}(E_{\mathrm{norm}}(C(u)))
\right],
\label{eq:sd}
\end{equation}
and uses $|D(u,C(u))|$ as the strength of prediction error. Here, $\mathrm{SD}=1$ indicates that a semantic inversion structure is present, while $|D|$ represents the degree of inversion.

Importantly, a large $|D|$ does not exclusively indicate sarcasm, but may also suggest other pragmatic phenomena such as exaggeration, humor, or deception. Therefore, this agent does not provide a sufficient condition for sarcasm, but is restricted to offering a candidate signal for a necessary condition in the form of semantic inversion. This restriction clarifies the necessity of downstream ToM reasoning by the Mental State and Intention Reasoner.

Overall, in this framework, this agent corresponds to prediction error computation in the world model.

\subsubsection{Mental state and intention reasoner}
\label{sec:intent_agent}

The Mental State and Intention Reasoner performs ToM-based reasoning on the internal mental states of the speaker, including emotion estimation and pragmatic intention inference, based on the input text $u$ and the estimated context $C(u)$. Its primary objective is to evaluate whether the utterance is likely to be produced with sarcastic intention. The output is represented as
\begin{equation}
T(u,C(u))=\{\text{intentions},\text{emotions}\},
\qquad
T_{\mathrm{sar}}(u,C(u))\in[0,1].
\end{equation}
Here, $T_{\mathrm{sar}}(u,C(u))$ is not a calibrated probability, but a ToM-based confidence score that indicates how well the utterance aligns with a typical sarcastic intention.

Sarcasm does not arise solely from the structural condition of semantic inversion, but is often accompanied by social and interpersonal intentions such as anger, disappointment, contempt, or distancing. By explicitly modeling this aspect through ToM reasoning, the Mental State and Intention Reasoner complements pragmatic distinctions that cannot be captured by structural signals such as $D$ and $\mathrm{SD}$ alone.

In addition to $T(u,C(u))$ and $T_{\mathrm{sar}}(u,C(u))$, the agent also generates an explanation $\rho_{\mathrm{intent}}(u,C(u))$ that provides the rationale behind the inference. The Mental State and Intention Reasoner takes the given context $C(u)$ as a fixed premise and does not reinterpret or revise it. Emotion estimation is treated as auxiliary evidence to support intention inference, and the final decision variable is $T_{\mathrm{sar}}$.

In summary, while the Inconsistency Detector evaluates the structural condition of whether literal evaluation and norm-based expectation are inverted, the Mental State and Intention Reasoner evaluates whether such an utterance is socially reasonable as a sarcastic intention. This division of roles enables the framework to distinguish cases where semantic inversion exists but the utterance is not sarcastic from cases that should be interpreted as sarcasm.

\subsection{Sarcasm arbitration with LR}
\label{subsec:arbiter}

The Sarcasm Arbiter is a decision layer that integrates the numerical summary signals produced by the agents and estimates the sarcasm probability of the input text $u$. This study does not delegate the final integration to an LLM, and instead estimates $P(\mathrm{sarcasm}=1\mid u)$ using a lightweight LR. LR can optimize data-driven weighting of numerical signals without performing any additional language reasoning, and its small number of parameters facilitates controlling overfitting. In addition, traceability of the reasoning basis is conducted at the module level by retaining the rationales generated by each agent, in addition to the LR coefficients themselves.

\subsubsection{Input features}
\label{sec:arbiter_features}

The LR input feature vector $\phi(u)$ uses the base features
\begin{equation}
\bigl(|D(u,C(u))|,\; T_{\mathrm{sar}}(u,C(u)),\; \mathrm{SD}(u,C(u))\bigr).
\end{equation}
Here, $|D|$ denotes the discrepancy strength between literal evaluation and norm-based expectation, $T_{\mathrm{sar}}$ denotes the ToM-based sarcasm intention alignment, and $\mathrm{SD}\in\{0,1\}$ denotes the presence or absence of sign disagreement.

In the implementation, feature engineering is additionally introduced not only with these three signals, but also for stabilizing the signals and representing interactions. Let $D:=|D|$ and $T:=T_{\mathrm{sar}}$. The following features are introduced:
\begin{align}
&D+T,\; D-T,\; T-D,\; D\times T,\;
\frac{D}{|T|+\epsilon},\; \frac{T}{|D|+\epsilon}, \\
&D^2,\; T^2,\; \sqrt{D},\; \sqrt{T},\;
\log(1+D),\; \log(1+T),
\end{align}
where $\epsilon$ is a small constant to avoid division by zero. In addition, sigmoid transformations using $\sigma(x)=1/(1+\exp(-x))$,
\begin{equation}
\sigma(D),\quad \sigma(T),
\end{equation}
are added as features. Furthermore, to reflect the structural condition given by $\mathrm{SD}$ in numerical features,
\begin{equation}
\mathrm{SD}\cdot D,\quad
\mathrm{SD}\cdot T,\quad
\mathrm{SD}\cdot(D+T),\quad
\mathrm{SD}\cdot(D-T)
\end{equation}
are included. Moreover, using the complement $\overline{\mathrm{SD}}=1-\mathrm{SD}$,
\begin{equation}
\overline{\mathrm{SD}},\quad
\overline{\mathrm{SD}}\cdot D,\quad
\overline{\mathrm{SD}}\cdot T
\end{equation}
are also added as features. This enables LR to learn how to handle signals in regions where sign disagreement does not hold. Any $\pm\infty$ or NaN that can arise during feature construction are replaced with 0 to ensure numerically safe inputs.

\subsubsection{LR formulation, model selection, and evaluation}
\label{sec:arbiter_lr}

LR outputs the sarcasm probability as follows:
\begin{equation}
P(\mathrm{sarcasm}=1\mid u)
=
\sigma\bigl(\mathbf{w}^{\top}\phi(u)+b\bigr),
\end{equation}
where $\sigma(\cdot)$ is the sigmoid function, and $\mathbf{w}$ and $b$ are trainable parameters. During training, standardization is applied to the input features, and class weights are balanced to address class imbalance. A regularization coefficient $C$ is introduced as a hyperparameter to control the inverse strength of L2 regularization. Specifically, stratified $K$-fold cross-validation is conducted on train+val obtained by concatenating train and val, and a grid search is performed over
\begin{equation}
C \in \{0.01,\,0.03,\,0.1,\,0.3,\,1.0,\,3.0,\,10.0\}.
\end{equation}
The objective for model selection is accuracy, and macro-F1 is used for tie-breaking when accuracy is identical.

The final label decision is not restricted to a fixed threshold of 0.5. For the predicted probabilities on the validation split in each CV fold, a threshold that maximizes accuracy is searched. The median of the optimal thresholds obtained across folds is adopted as the final threshold $\tau$:
\begin{equation}
\hat{y}=\mathbb{I}\bigl[P(\mathrm{sarcasm}=1\mid u)\ge \tau\bigr].
\end{equation}
This enables data-driven adjustment of the decision criterion against distributional differences and class imbalance across datasets.

The final model is retrained on the entire train+val under the selected $(C,\tau)$ and evaluated on the test split. Evaluation metrics include accuracy and macro-F1 for the binary classification.

\subsection{Algorithm and world model interpretation}
\label{sec:algo_wm}

This subsection presents the inference procedure of WM-SAR in Algorithm~\ref{alg:sarcasm_lr} and provides an interpretation as a world model computational structure. WM-SAR consists of three stages: LLM agents extract cognitive signals such as literal evaluation, latent context, norm-based expectation, and intention alignment as scalars; the discrepancy between observation and prediction is deterministically computed as a difference; and the resulting low-dimensional signals are integrated by a lightweight LR to output a sarcasm probability and a label. A key point is that the components corresponding to prediction error are not delegated to free-form LLM reasoning, but are explicitly isolated as $D(u,C(u))$ and $\mathrm{SD}(u,C(u))$, thereby enabling the core structure of semantic inversion to be directly handled as reproducible numerical quantities.

\begin{algorithm}[t]
\caption{World model inspired sarcasm reasoning with LLM agents and LR}
\label{alg:sarcasm_lr}
\begin{algorithmic}[1]
\Require Utterance $u$, trained agent set $\mathcal{A}=\{A_{\mathrm{literal}},A_{\mathrm{context}},A_{\mathrm{norm}},A_{\mathrm{inc}},A_{\mathrm{intent}}\}$, 
trained LR parameters $(\mathbf{w}, b)$, decision threshold $\tau$
\Ensure Sarcasm probability $P(\mathrm{sarcasm}=1\mid u)$ and predicted label $\hat{y}$
\Statex
\Comment{Agent-based signal extraction}
\State $(M_{\mathrm{literal}}(u), \rho_{\mathrm{literal}}) \gets A_{\mathrm{literal}}(u)$
\State $(C(u), \rho_{\mathrm{context}}) \gets A_{\mathrm{context}}(u)$
\State $(E_{\mathrm{norm}}(C(u)), \rho_{\mathrm{norm}}) \gets A_{\mathrm{norm}}(C(u))$
\State $(T_{\mathrm{sar}}(u,C(u)), \rho_{\mathrm{intent}}) \gets A_{\mathrm{intent}}(u,C(u))$
\Statex
\Comment{Deterministic prediction error computation}
\State $D(u,C(u)) \gets M_{\mathrm{literal}}(u) - E_{\mathrm{norm}}(C(u))$
\State $\mathrm{SD}(u,C(u)) \gets \mathbb{I}\!\left[\operatorname{sgn}(M_{\mathrm{literal}}(u)) \neq \operatorname{sgn}(E_{\mathrm{norm}}(C(u)))\right]$
\Statex
\Comment{Feature construction for LR arbiter}
\State $D \gets |D(u,C(u))|,\quad T \gets T_{\mathrm{sar}}(u,C(u)),\quad S \gets \mathrm{SD}(u,C(u))$
\State Construct base feature vector $\phi_{\text{base}}(u) \gets (D,\; T,\; S)$
\State Construct engineered features (examples):
\State \hspace{1em} $D+T,\; D-T,\; T-D,\; D\times T,\; \frac{D}{|T|+\epsilon},\; \frac{T}{|D|+\epsilon}$
\State \hspace{1em} $D^2,\; T^2,\; \sqrt{D},\; \sqrt{T},\; \log(1+D),\; \log(1+T),\; \sigma(D),\; \sigma(T)$
\State \hspace{1em} $S\cdot D,\; S\cdot T,\; S\cdot(D+T),\; S\cdot(D-T),\; (1-S),\; (1-S)\cdot D,\; (1-S)\cdot T$
\State $\phi(u) \gets$ concatenation of $\phi_{\text{base}}(u)$ and engineered features
\Statex
\Comment{Sarcasm arbitration by LR}
\State $P(\mathrm{sarcasm}=1\mid u) \gets \sigma\!\left(\mathbf{w}^{\top}\phi(u) + b\right)$
\State $\hat{y} \gets \mathbb{I}\!\left[P(\mathrm{sarcasm}=1\mid u) \ge \tau\right]$
\State \Return $P(\mathrm{sarcasm}=1\mid u),\; \hat{y}$
\end{algorithmic}
\end{algorithm}

Next, this algorithm is interpreted by mapping it to typical components of a world model. WM-SAR does not introduce a reinforcement learning style world model with environment transitions, rewards, or policy learning. Instead, it focuses on reconstructing the computational structure observation $\rightarrow$ latent state $\rightarrow$ prediction $\rightarrow$ prediction error $\rightarrow$ decision through role specialization of LLM agents and deterministic computation. Since each module has a corresponding scalar output, module-wise ablation and contribution analysis are possible, enabling verification of the reasoning process at the level of computational elements. The correspondence is summarized in Table~\ref{tab:world_model_mapping}.

\begin{table}[t]
\centering
\caption{Correspondence between world model components and modules in the WM-SAR}
\label{tab:world_model_mapping}
\begin{tabular}{ll}
\hline
World model component & Corresponding module in this work \\
\hline
Observation & Literal Agent ($M_{\mathrm{literal}}$) \\
Latent state inference & Context Agent ($C(u)$) \\
State prediction & Norm Agent ($E_{\mathrm{norm}}$) \\
Prediction error & Inconsistency Detector ($D$, $\mathrm{SD}$) \\
Decision & Intent Agent ($T_{\mathrm{sar}}$) + LR Arbiter \\
\hline
\end{tabular}
\end{table}

Overall, WM-SAR reinterprets a world model not as an internal representation to be learned, but as a verifiable reasoning structure, and is positioned as a framework that separates and integrates knowledge-driven reasoning by LLMs with deterministic difference computation and lightweight learning.

\section{Experiment and analysis}\label{sec4}
\subsection{Experiment design}\label{subsec12}

\subsubsection{Dataset}
\label{subsubsec10}

This study conducts evaluation using three representative sarcasm detection datasets with different domains and language usage characteristics.

\begin{itemize}
\item[i)] IAC-V1 \cite{30}:  
A dataset composed of comments collected from an online political discussion forum, containing many sarcastic expressions embedded in argumentative contexts. Many utterances involve explicit claim structures and dialogic backgrounds, requiring pragmatic reasoning ability.

\item[ii)] IAC-V2 \cite{31}:  
A large-scale dataset that extends IAC-V1 and includes more diverse speakers and topics. With an increased number of sarcastic and non-sarcastic instances and higher lexical and syntactic diversity, it is suitable for examining model generalization performance.

\item[iii)] SemEval-2018 Task 3 \cite{32}:  
A sarcasm and irony detection dataset for English tweets on Twitter. Since it contains short and informal expressions including slang and abbreviated forms, implicit intention reasoning is required under scarce explicit context.
\end{itemize}

These datasets enable comprehensive evaluation across different language phenomena, including sarcasm understanding in argumentative long-form texts and sarcasm understanding in short and colloquial utterances. For each dataset, the data are split into training, validation, and test sets following the ratio 0.8, 0.1, and 0.1, and training, validation, and testing are conducted accordingly.

\subsubsection{Baseline methods}
\label{subsubsec13}

To verify the effectiveness of the proposed method, comparisons are conducted with the following representative baselines, ranging from conventional deep learning models to LLM-based reasoning approaches.

\begin{itemize}
\item[i)] MIARN \cite{33}:  
A multi-stage attentive neural model that captures sarcasm expressions by introducing an intra-attention mechanism and emphasizing important words and cues in the utterance.

\item[ii)] SAWS \cite{34}:  
A model that focuses on polarity inversion and opposition in sentiment and emphasizes sarcasm-indicative parts through a weighted attention mechanism.

\item[iii)] DC-Net \cite{35}:  
A deep learning model with a dual-channel architecture that explicitly models the interrelationship between an utterance and its context.

\item[iv)] BERT \cite{36}:  
A standard baseline obtained by fine-tuning a pre-trained language model for the sarcasm detection task, providing discriminative performance based on contextual representations.

\item[v)] GPT-4.1-mini:  
A baseline that performs sarcasm judgment by zero-shot reasoning with a LLM, without any additional training.

\item[vi)] GPT-4.1-mini + CoC \cite{24}:  
A method using a prompting strategy based on Chain of Contradiction (CoC) to conduct step-by-step reasoning over semantic contradictions in the utterance.

\item[vii)] GPT-4.1-mini + GoC \cite{24}:  
A method that reasons by structuring multiple sarcasm cues and associating them through Graph of Cues (GoC).

\item[viii)] GPT-4.1-mini + BoC \cite{24}:  
An ensemble-style method that aggregates multiple cue-based reasoning outputs to make the final judgment through Bagging of Cues (BoC).

\item[ix)] CAF-I \cite{7}:  
A collaborative reasoning framework in which multiple LLM agents interpret the utterance from different perspectives and integrate their outputs to judge sarcasm.
\end{itemize}

In this study, GPT-4.1-mini is used for all agent backbone models in the proposed method, as well as for the LLM-based baselines. This enables fair comparison of how simple zero-shot reasoning, existing prompt-engineering strategies, and the proposed world model inspired reasoning structure affect performance.

\subsubsection{Evaluation on performance comparison}\label{subsubsec14}

This subsection evaluates the sarcasm detection performance of the proposed WM-SAR in comparison with deep learning models and LLM-based methods. The evaluation is conducted on three datasets, IAC-V1, IAC-V2, and SemEval-2018, using Accuracy and Macro-F1 as metrics. For WM-SAR, the LR regularization coefficient $C$ and the decision threshold $\tau$ are optimized for each dataset via cross-validation on train+val. The comparison results are shown in Table~\ref{tab:perf_comp}.

WM-SAR achieves the best performance on all three datasets in terms of both Accuracy and Macro-F1. In particular, WM-SAR consistently outperforms the simple zero-shot GPT-4.1-mini baseline as well as advanced prompting approaches such as CoC, GoC, and BoC. This indicates that, rather than directly using the raw output of an LLM, decomposing and integrating the reasoning process based on a world model structure is an effective design for sarcasm detection.

WM-SAR also demonstrates consistently strong performance across all datasets compared with task-specific deep learning methods such as MIARN, SAWS, and DC-Net, as well as the fine-tuned BERT baseline. This suggests that leveraging an LLM with large-scale pretraining knowledge as cognitive modules, while integrating their outputs through deterministic difference computation and a lightweight learner, enables high generalization ability under different data distributions and context diversity.

Overall, WM-SAR consistently outperforms existing deep learning models and LLM-based methods across diverse sarcasm detection benchmarks, empirically demonstrating the effectiveness of the proposed world model inspired reasoning framework.

\begin{table*}[t]
\centering
\caption{Performance comparison with baseline methods}
\label{tab:perf_comp}
\begin{tabular}{l|cc|cc|cc|cc}
\hline
Method & \multicolumn{2}{c|}{IAC-V1} & \multicolumn{2}{c|}{IAC-V2} & \multicolumn{2}{c|}{SemEval-2018} & \multicolumn{2}{c}{Avg.} \\
 & Acc. & F1 & Acc. & F1 & Acc. & F1 & Acc. & F1 \\
\hline
MIARN & 0.650 & 0.641 & 0.736 & 0.735 & 0.652 & 0.650 & 0.679 & 0.675 \\
SAWS & 0.480 & 0.476 & 0.599 & 0.598 & 0.610 & 0.600 & 0.563 & 0.558 \\
DC-Net & 0.570 & 0.557 & 0.714 & 0.713 & 0.675 & 0.675 & 0.653 & 0.648 \\
BERT & 0.655 & 0.655 & 0.791 & 0.790 & 0.707 & 0.694 & 0.718 & 0.713 \\
GPT-4.1-mini & 0.725 & 0.724 & 0.768 & 0.768 & 0.699 & 0.698 & 0.731 & 0.730 \\
GPT-4.1-mini+CoC & 0.685 & 0.685 & 0.776 & 0.775 & 0.693 & 0.676 & 0.718 & 0.712 \\
GPT-4.1-mini+GoC & 0.685 & 0.684 & 0.763 & 0.759 & 0.660 & 0.650 & 0.702 & 0.698 \\
GPT-4.1-mini+BoC & 0.695 & 0.695 & 0.773 & 0.773 & 0.699 & 0.697 & 0.722 & 0.722 \\
CAF-I & 0.665 & 0.644 & 0.672 & 0.648 & 0.657 & 0.638 & 0.665 & 0.643 \\
\hline
WM-SAR (Ours) & \textbf{0.745} & \textbf{0.745} & \textbf{0.791} & \textbf{0.791} & \textbf{0.714} & \textbf{0.714} & \textbf{0.750} & \textbf{0.750} \\
\hline
\end{tabular}
\end{table*}
\subsection{Ablation study}\label{subsubsec15}

\begin{figure}[t]
    \centering
    \includegraphics[width=\linewidth]{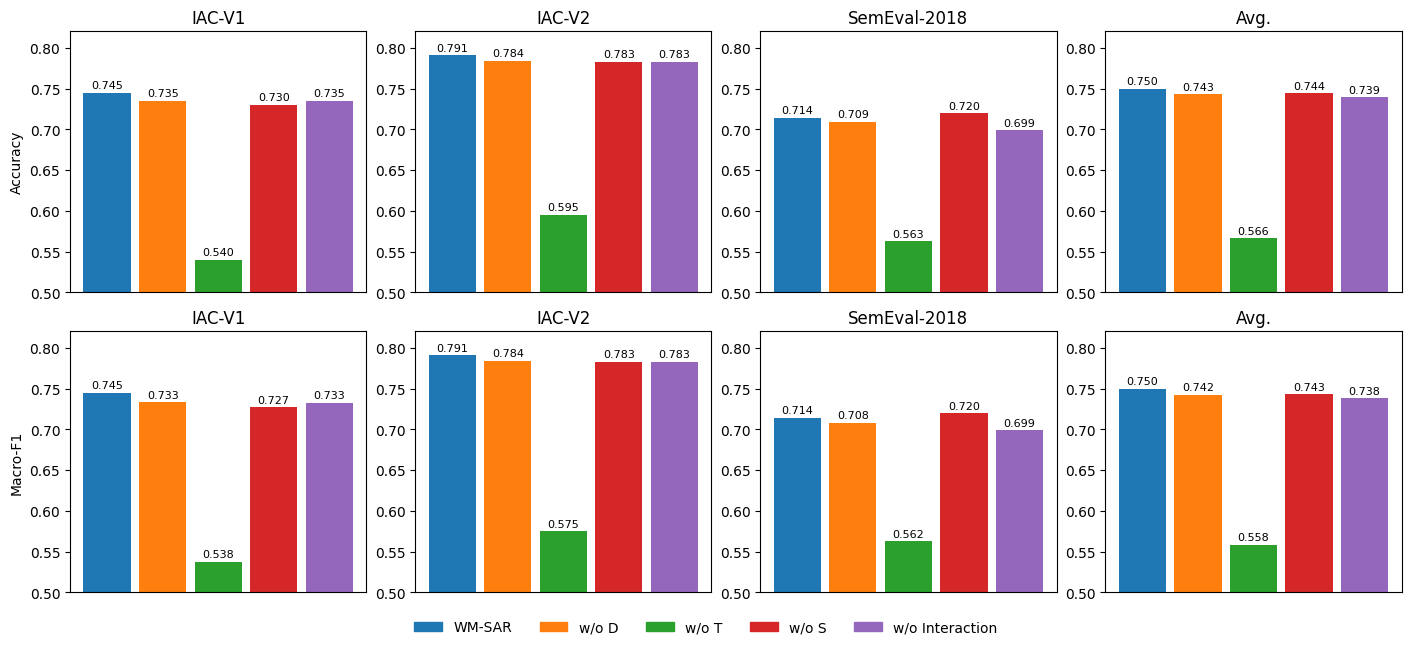}
    \caption{Ablation study of WM-SAR.}
    \label{fig:ablation}
\end{figure}

This subsection conducts ablation experiments to analyze the contributions of the main numerical signals and feature design in WM-SAR. Specifically, the following settings are compared with the full WM-SAR: removing the inconsistency strength $|D|$ (w/o D), removing the ToM-based sarcasm intention score $T_{\mathrm{sar}}$ (w/o T), removing the sign disagreement indicator $\mathrm{SD}$ (w/o S), and removing interaction features by using only the three base signals (w/o Interaction). For each setting, Accuracy and Macro-F1 are evaluated on IAC-V1, IAC-V2, SemEval-2018, and their average. The results are shown in the table and Fig.~\ref{fig:ablation}.

First, w/o T, which removes $T_{\mathrm{sar}}$, shows a substantial performance drop across all datasets, confirming that the ToM-based intention reasoning signal is a core component of WM-SAR. This is consistent with the design principle of this study: semantic inversion structure alone is insufficient to reliably identify sarcasm, and estimating the speaker's social intention and emotional state is indispensable.

Next, w/o D, which removes the inconsistency strength $|D|$, yields a relatively smaller but consistently observed performance decrease, indicating that the discrepancy magnitude between literal evaluation and norm-based expectation serves as an effective structural signal that complements the strength of sarcasm. Similarly, w/o S, which removes the sign disagreement indicator $\mathrm{SD}$, also reduces performance, suggesting that explicitly modeling the presence of semantic inversion contributes to discrimination performance. However, on SemEval-2018, w/o S achieves performance close to the full model, implying that in short and noisy SNS-domain utterances, $|D|$ and $T_{\mathrm{sar}}$ may play more dominant roles.

In addition, w/o Interaction, which removes interaction features, shows an overall performance decrease relative to the full model, confirming that capturing non-linear relationships among $|D|$, $T_{\mathrm{sar}}$, and $\mathrm{SD}$ through LR is beneficial for improving final decision accuracy. This supports the goal of the proposed method: sarcasm cues arise not only from individual signals but also from combinations of multiple world model-derived signals.

Overall, the performance gains of WM-SAR are achieved through complementary functioning of ToM-based intention reasoning, inconsistency strength representing the discrepancy between observation and expectation, the structural condition indicating semantic inversion, and the representation of their interactions. These ablation results support that the proposed world model-style decomposition is not merely using LLM outputs, but a design that integrates each cognitive component as an explicit signal.

\subsection{Evaluation on model robustness}\label{subsec16}

This subsection evaluates the robustness of WM-SAR with respect to the backbone model used for the LLM agents. Specifically, each agent is instantiated with one of three models, GPT-4.1, GPT-4.1-mini, and GPT-4.1-nano, while keeping all other settings identical, and the resulting performance is compared. The results are evaluated using Accuracy and Macro-F1 on IAC-V1, IAC-V2, and SemEval-2018, as well as their averages, and are reported in Table~\ref{tab:robustness}.

\begin{table}[t]
\centering
\caption{Robustness of WM-SAR with different LLM backbones.}
\label{tab:robustness}
\begin{tabular}{lcccccccc}
\hline
 & \multicolumn{2}{c}{IAC-V1} & \multicolumn{2}{c}{IAC-V2} & \multicolumn{2}{c}{SemEval-2018} & \multicolumn{2}{c}{Avg.} \\
Method & Acc. & F1 & Acc. & F1 & Acc. & F1 & Acc. & F1 \\
\hline
GPT-4.1       & 0.725 & 0.725 & \textbf{0.792} & \textbf{0.792} & \textbf{0.736} & \textbf{0.735} & \textbf{0.751} & \textbf{0.751} \\
GPT-4.1-mini  & \textbf{0.745} & \textbf{0.745} & 0.791 & 0.791 & 0.714 & 0.714 & 0.750 & 0.750 \\
GPT-4.1-nano  & 0.645 & 0.644 & 0.706 & 0.702 & 0.607 & 0.590 & 0.653 & 0.646 \\
\hline
\end{tabular}
\end{table}

WM-SAR with GPT-4.1-mini exhibits performance that is nearly comparable to that obtained with the full-size GPT-4.1, with only a negligible difference in the average results. This suggests that WM-SAR does not strongly depend on the final decision capability of the LLM itself, but rather operates as a framework that structures and integrates intermediate reasoning signals extracted from the LLM, including literal evaluation, a context hypothesis, norm-based expectation, and intention reasoning. In other words, once the backbone model reaches a sufficient level of reasoning ability, even a lightweight model can effectively support this framework.

In contrast, when using GPT-4.1-nano, a clear performance degradation is observed across all datasets. This is likely because extremely small models provide insufficient quality in complex social reasoning, including context construction, norm-based expectation inference, and ToM-based intention reasoning, which in turn degrades the quality of the numerical signals supplied to WM-SAR. Nevertheless, WM-SAR still operates stably with the nano model and does not exhibit catastrophic failure, which indicates structural robustness of the proposed method.

Overall, WM-SAR is relatively robust to the choice of the backbone LLM. In particular, even when using a lightweight model such as GPT-4.1-mini, WM-SAR maintains accuracy comparable to that achieved with a higher-capacity model. This property implies that the proposed framework can be realistically applied in environments with computational resource or cost constraints, and supports the practical utility of WM-SAR.
\subsection{Evaluation on computational cost}\label{subsec17}

This subsection evaluates the computational cost of WM-SAR from the perspective of efficiency in practical deployment by comparing it with existing methods. For each method, the mean (Mean) and standard deviation (STD) of inference time per sample are measured, and the results are reported in Table~\ref{tab:cost}.

\begin{table}[t]
\centering
\caption{Computational cost comparison in terms of inference time per sample.}
\label{tab:cost}
\begin{tabular}{lcc}
\hline
Method & Mean & STD \\
\hline
GPT-4.1-mini         & 0.8987 & 0.1677 \\
GPT-4.1-mini+CoC     & 0.7667 & 0.0272 \\
GPT-4.1-mini+GoC     & 1.2427 & 0.1560 \\
GPT-4.1-mini+BoC     & 1.0027 & 0.1209 \\
CAF-I                & 21.1047 & 1.0605 \\
WM-SAR               & 7.6520 & 0.6622 \\
\hline
\end{tabular}
\end{table}

Direct inference with a single GPT-4.1-mini as well as prompt-extended methods such as CoC, GoC, and BoC all require roughly around one second per sample. This reflects their configurations, which call an LLM once or only a small number of times, and therefore remain relatively low-cost.

In contrast, CAF-I, which performs multi-stage reasoning with multiple agents and cross-referencing among them, requires an extremely large computational cost, exceeding 20 seconds on average. While it provides high accuracy, this cost represents a substantial burden for practical deployment.

WM-SAR, which also uses multiple agents, requires more inference time than single-prompt methods, but the mean inference time is kept to approximately 7.65 seconds, which is substantially lower than that of CAF-I. This reduction is attributed to the design of WM-SAR: LLM outputs are extracted once as numerical signals, and the final decision is performed by a lightweight LR, thereby avoiding unnecessary additional reasoning or iterative LLM calls.

Overall, WM-SAR requires more computation than simple LLM-based methods, but its computational cost remains practical for a high-accuracy method that introduces a world model inspired reasoning structure. In particular, compared with CAF-I, WM-SAR achieves substantial speedup while maintaining competitive performance, indicating that WM-SAR provides a favorable balance between accuracy and efficiency.
\subsection{Evaluation on explainability}
\label{subsec18}

This subsection evaluates how understandable the decision process of WM-SAR is for humans. Based on the intermediate scalar signals output by each agent, $M_{\mathrm{literal}}$, $E_{\mathrm{norm}}$, $|D|$, $T_{\mathrm{sar}}$, and $\mathrm{SD}$, together with their associated rationales, interpretability analysis is conducted from three perspectives: statistical comparison between correctly and incorrectly classified samples, contribution analysis of the final decision through LR weight interpretation, and qualitative analysis via representative case studies.

\subsubsection{Statistical analysis of intermediate signals}

First, Table~\ref{tab:explain_stats} reports the comparison of the mean values of the main intermediate signals and the rate of sign disagreement between correctly classified and misclassified samples in the test set.

\begin{table}[t]
\centering
\caption{Statistics of intermediate signals for correctly and incorrectly classified samples.}
\label{tab:explain_stats}
\begin{tabular}{lcc}
\hline
Metric & Correct & Wrong \\
\hline
$T_{\mathrm{sar}}$ (mean) & 0.417 & 0.394 \\
$|D|$ (mean)             & 0.432 & 0.439 \\
$\mathrm{SD}$ rate       & 0.245 & 0.286 \\
\hline
\end{tabular}
\end{table}

The results indicate that there is no large discrepancy in the mean values of $T_{\mathrm{sar}}$ or $|D|$ between correct and incorrect samples, suggesting that sarcasm cannot be clearly separated by any single signal alone. In contrast, the rate of $\mathrm{SD}$ is slightly higher for misclassified samples, implying that the sign relation between literal meaning and norm-based expectation alone is insufficient for reliable discrimination. These observations support the necessity of integrating multiple signals in WM-SAR.

\subsubsection{Interpretation of LR weights}
\label{sec:lr_weight_analysis}

\begin{figure}[t]
    \centering
    \includegraphics[width=\linewidth]{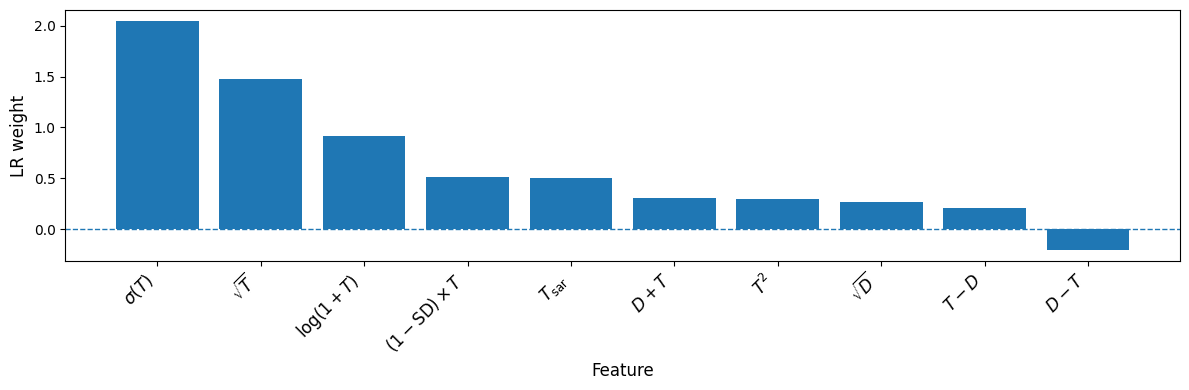}
    \caption{
    Top-10 ranked LR weights for the WM-SAR arbiter.
    }
    \label{fig:lr_weights}
\end{figure}

Next, the weights of the LR used as the Sarcasm Arbiter are analyzed to examine how much each feature contributes to the final decision. Fig.~\ref{fig:lr_weights} shows the top-10 features with the largest absolute LR coefficients after inverse standardization. Positive weights indicate contribution to the sarcasm class, while negative weights indicate contribution to the non-sarcasm class.

The largest positive weights correspond to features derived from the ToM-based sarcasm intention score, such as $\sigma(T_{\mathrm{sar}})$, $\sqrt{T_{\mathrm{sar}}}$, and $\log(1+T_{\mathrm{sar}})$, and $T_{\mathrm{sar}}$ itself also exhibits a high contribution. This result indicates that the intention reasoning signal plays a dominant role in the final decision, which is consistent with the substantial performance drop observed in the w/o $T$ ablation.

In addition, interaction and nonlinear features such as $(1-\mathrm{SD})\!\times\!T_{\mathrm{sar}}$, $D+T_{\mathrm{sar}}$, and $T_{\mathrm{sar}}^2$ also appear among the top features, showing that the intention signal contributes to sarcasm judgment in combination with semantic inconsistency and the structural condition. This suggests that the final decision is formed by combinations of multiple world model inspired signals rather than by any single signal.

On the other hand, $\sqrt{|D|}$ also has a positive contribution, confirming that the discrepancy strength between literal meaning and norm-based expectation functions as a supporting structural signal for sarcasm. Furthermore, $T_{\mathrm{sar}}-D$ has a positive weight, while $D-T_{\mathrm{sar}}$ has a negative weight, reflecting the design principle of WM-SAR: sarcasm is more likely when the intention signal outweighs the discrepancy, whereas a large discrepancy alone without strong intention is less likely to be judged as sarcasm.

Overall, the LR weight analysis quantitatively supports that ToM-based intention reasoning is the core component, semantic inconsistency acts as an auxiliary structural signal, and their interactions are crucial for the final decision. This result confirms at the decision level that WM-SAR integrates interpretable signals in a transparent manner.

\subsubsection{Case study: correctly classified example}

An example that is correctly classified as non-sarcasm is shown below.

\begin{quote}
\small
\textit{This is the line that caused me to butt in, walking-fish. Again, my apologies, but prs has no idea of the difference between Lamarckian and Darwinian theory, and how that theoretical difference would be reflected in different social theories. As far as I can tell prs has nothing but preconceived misconceptions about this. I felt compelled to speak up. Hope you're not put out about my intrusion.}
\end{quote}

In this case, the Literal Meaning Agent focuses on apologetic expressions and a polite tone, assigning a weak positive polarity of $M_{\mathrm{literal}} = 0.10$. The Norm and Expectation Reasoner judges that correcting a misunderstanding and constructively intervening in a discussion is socially desirable, estimating a relatively high norm-based expectation of $E_{\mathrm{norm}} = 0.60$. As a result, although $|D| = 0.50$ indicates a certain discrepancy, the signs agree and thus $\mathrm{SD}=0$. Furthermore, the Mental State \& Intention Reasoner detects almost no sarcasm intention and outputs a low value of $T_{\mathrm{sar}} = 0.10$.

By integrating these signals, the model interprets this utterance as a sincere corrective action and correctly predicts non-sarcasm. This example shows that when all agents provide consistent interpretations, WM-SAR produces decisions aligned with human intuition.

\subsubsection{Case study: misclassified example}

Next, an example that is sarcastic but is misclassified as non-sarcasm is shown.

\begin{quote}
\small
\textit{A truly charming publication. Overflowing with the warmth of human goodness.}
\end{quote}

This utterance is composed of superficially highly positive expressions. The Literal Meaning Agent assigns a strong positive polarity of $M_{\mathrm{literal}} = 0.80$, and the Norm and Expectation Reasoner also regards praise as socially desirable, giving $E_{\mathrm{norm}} = 0.80$. Consequently, $|D| = 0.00$ indicates no discrepancy between meaning and norm, and $\mathrm{SD}=0$. Moreover, the Mental State \& Intention Reasoner fails to detect sarcasm intention and outputs a low value of $T_{\mathrm{sar}} = 0.10$.

Since all agents interpret the utterance consistently as sincere praise, WM-SAR fails to detect sarcasm. This example represents a typical case of so-called positive sarcasm, in which meaning, norm, and intention are all superficially aligned, highlighting an inherent difficulty of the proposed method.

From these analyses, WM-SAR is shown to achieve both high performance and high interpretability when a discrepancy between meaning and norm exists and the Mental State \& Intention Reasoner detects sarcasm intention. In contrast, when all signals are superficially consistent, misclassification is more likely to occur.

Nevertheless, by jointly inspecting the outputs and rationales of each agent together with the LR weights, WM-SAR enables humans to trace which signals contributed to the decision and at which stage a misunderstanding occurred. In this respect, unlike a single black-box classifier, the proposed method provides a major advantage in that its decision rationale can be explained in a linguistic, structural, and quantitative manner.

\section{Discussion}
\label{sec5}

\subsection{Key findings}

The key finding of this study is that, rather than using an LLM as a single black-box classifier, decomposing sarcasm understanding into a world model inspired structure of meaning, norm, intention, and inconsistency, explicitly modeling them as independent agents and integrating their outputs, enables the simultaneous achievement of high performance and interpretability.

From the performance perspective, WM-SAR consistently outperforms existing deep learning methods, fine-tuned PLMs, as well as standalone LLMs and advanced prompting-based approaches on IAC-V1, IAC-V2, and SemEval-2018. In particular, the fact that structured reasoning yields consistent improvements even compared with GPT-4.1-mini alone, various prompting strategies, and multi-agent baselines suggests that better reasoning structure, rather than simply stronger LLMs, is essential for sarcasm understanding.

The ablation study shows that $|D|$, $T_{\mathrm{sar}}$, and their interactions all contribute to performance, with a substantial degradation observed when $T_{\mathrm{sar}}$ is removed. This result quantitatively supports that sarcasm understanding is not merely polarity reversal detection, but inherently involves a ToM-based component of speaker intention reasoning.

Furthermore, the robustness analysis demonstrates that the relative trends of WM-SAR remain consistent even when different sizes of LLMs (GPT-4.1, mini, nano) are used, indicating that the core structural design is less sensitive to model scale. This suggests that the proposed framework is not a heuristic tightly coupled to a specific LLM, but can function as a more general reasoning structure.

From the perspective of computational cost, WM-SAR is substantially more efficient than multi-stage and recursive LLM reasoning frameworks such as CAF-I. Although it incurs higher cost than simple LLM + prompting approaches, it remains within a practical range given the achieved performance gains. This confirms the effectiveness of combining deterministic computation with a lightweight learner (LR), rather than delegating everything to LLM inference.

The interpretability analysis and case studies further show that, even when WM-SAR makes errors, humans can trace which agents produced what interpretations and which signals were missing or malfunctioned. This property is qualitatively different from conventional black-box classifiers, as it enables structural explanations of why a particular decision was made for an inherently ambiguous and pragmatic phenomenon such as sarcasm.

Taken together, this study demonstrates that sarcasm understanding can be reinterpreted as a world model style reasoning structure of observation $\rightarrow$ latent state $\rightarrow$ prediction $\rightarrow$ prediction error $\rightarrow$ decision, and that implementing this structure through LLM-based linguistic reasoning and deterministic computation achieves a favorable balance among performance, efficiency, and interpretability. This perspective is not limited to sarcasm, but also suggests that other NLP tasks involving implicature, euphemism, humor, offensive language, and more generally social reasoning and value judgment, may benefit from being reformulated as world model inspired structures to more systematically exploit LLM capabilities. In this sense, this work argues that, in the LLM era, the design of reasoning structures, rather than mere model scaling, is a crucial direction for task design.

\subsection{Limitations}

This study has three main limitations.

First, the LLMs used in this work are mainly limited to the GPT-4.1 family, and we do not evaluate WM-SAR with models based on different architectures or pretraining strategies. Therefore, the extent to which the effectiveness of WM-SAR is model-agnostic remains to be verified, and future work should conduct comparative experiments with a more diverse set of LLMs.

Second, our evaluation is based on sentence-level sarcasm datasets and does not explicitly consider sarcasm understanding that requires dialogue history, discourse structure, or speaker relationships. As a result, the effectiveness of the Context Constructor Agent for sarcasm phenomena that rely on multi-turn interactions or long-term context remains untested. Extending the framework to context-enriched datasets is an important direction for future research.

Third, since WM-SAR is designed around the discrepancy between meaning and norm and ToM-based intention reasoning, it is inherently prone to misclassification in so-called positive sarcasm, where surface meaning and norm-based expectation align and sarcasm intention is difficult to detect from context. This limitation stems from the inductive bias of the proposed framework, and future improvements may be achieved by incorporating more subtle pragmatic cues or discourse-level signals.


\section{Conclusion}
\label{sec6}

In this study, we address sarcasm understanding, one of the most challenging problems in NLP, by focusing on its essential property: the discrepancy between surface lexical polarity and the speaker’s latent intention and context, and by investigating how such a cognitive structure can be modeled and handled in an interpretable manner. While existing deep learning and LLM-based methods achieve strong performance, they tend to operate as black boxes, making it difficult to structurally explain why a particular sarcasm judgment is made. To tackle this issue, we present a new perspective that reinterprets sarcasm understanding as a world model style reasoning process.

Specifically, we propose the WM-SAR framework, which explicitly models literal meaning, context construction, norm-based expectation, and speaker intention as role-specialized LLM agents, deterministically computes inconsistency from their outputs, and integrates the final decision using a lightweight LR. The framework reconstructs the reasoning process in correspondence with the world model structure of observation $\rightarrow$ latent state $\rightarrow$ prediction $\rightarrow$ prediction error $\rightarrow$ decision, and is characterized by separating and integrating LLM-based linguistic reasoning with numerical and structural computation. This design enables the explicit handling of which cognitive discrepancies are decisive in sarcasm judgment.

Through experiments on three benchmarks, IAC-V1, IAC-V2, and SemEval-2018, we show that WM-SAR consistently outperforms strong baselines, including conventional deep learning models and GPT-based standalone and advanced prompting approaches. In particular, ablation studies confirm that integrating the semantic inconsistency magnitude $|D|$ and the sarcasm intention score $T_{\mathrm{sar}}$ is essential for performance gains. Moreover, robustness evaluations across different LLM sizes and computational cost analysis demonstrate that WM-SAR achieves a favorable balance between accuracy and efficiency. In addition, case studies based on agent rationales and intermediate features show that both correct and erroneous predictions can be analyzed in a human-traceable manner, thereby empirically validating the interpretability of the decision process that is difficult to obtain with black-box approaches.

Future work includes validating generality with different types of LLMs, extending the framework to context-dependent sarcasm datasets that include dialogue history, and developing toward multimodal information and interactive world models. Through these directions, we expect that the world model inspired view of sarcasm understanding presented in this study can be further extended to a broader range of pragmatic phenomena and social reasoning tasks.


\bmhead{Supplementary information}

Supplementary information is not available.

\bmhead{Acknowledgements}

The authors would like to thank The Nippon Foundation HUMAI Program for supporting this study and for providing a research environment that enabled the completion of this work.

\section*{Declarations}

\begin{itemize}
\item \textbf{Funding} \\
This work was supported by The Nippon Foundation HUMAI Program.

\item \textbf{Conflict of interest/Competing interests} \\
The authors declare that they have no competing interests.

\item \textbf{Ethics approval and consent to participate} \\
Not applicable. This study used only publicly available datasets and did not involve any human participants or personal data.

\item \textbf{Consent for publication} \\
Not applicable.

\item \textbf{Data availability} \\
The datasets used in this study are publicly available from their official sources, including IAC-V1, IAC-V2, and SemEval-2018 Task 3.

\item \textbf{Materials availability} \\
Not applicable.

\item \textbf{Code availability} \\
The code will be publicly released upon acceptance of this manuscript.

\item \textbf{Author contribution} \\
Keito Inoshita conceived the study, designed the methodology, implemented the model, conducted the experiments, and wrote the manuscript.  
Shinnosuke Mizuno contributed to the research design and provided conceptual guidance.

\end{itemize}

\bibliography{sn-bibliography}

\end{document}